\documentclass[conference]{IEEEtran}
\IEEEoverridecommandlockouts
\usepackage{cite}
\usepackage{amsmath,amssymb,amsfonts}
\usepackage{algorithmic}
\usepackage{graphicx}
\usepackage{textcomp}
\usepackage{xcolor}
\usepackage{hyperref}

\begin{document}
\title{Drone-type-Set: Drone types detection benchmark for drone detection and tracking}
\author{Kholoud AlDosari, AIbtisam Osman, Omar Elharrouss, Somaya Al\-Maadeed, Mohamed Zied Chaari}
\maketitle
\begin{abstract}
The Unmanned Aerial Vehicles (UAVs) market has been significantly growing and Considering the availability of drones at low-cost prices the possibility of misusing them, for illegal purposes such as drug trafficking, spying, and terrorist attacks posing high risks to national security, is rising. Therefore, detecting and tracking unauthorized drones to prevent future attacks that threaten lives, facilities, and security, become a necessity. Drone detection can be performed using different sensors, while image-based detection is one of them due to the development of artificial intelligence techniques. However, knowing unauthorized drone types is one of the challenges due to the lack of drone types datasets. For that, in this paper, we provide a dataset of various drones as well as a comparison of recognized object detection models on the proposed dataset including YOLO algorithms with their different versions, like, v3, v4, and v5 along with the Detectronv2. The experimental results of different models are provided along with a description of each method. The collected dataset can be found in \url{https://drive.google.com/drive/folders/1EPOpqlF4vG7hp4MYnfAecVOsdQ2JwBEd?usp=share_link}    
\end{abstract}


\section{Introduction}

Recently, Unmanned Aerial Vehicles (UAVs) are being widely used in various sectors, the diversity of drones along with their affordable prices make them attractive to use. Drones are engaged in applications such as health care, weather forecasting, and traffic surveillance \cite{b1}. Many people use the new technology of aerial vehicles as toys or to help them accomplish their daily tasks. Moreover, drones can be modified to serve the needs \cite{b2}. For surveillance and security purposes, the drones can map the fires or help to evacuate crowded places. Recreational drones are used for cinematography or for other entertaining purposes. However, the easy access of drones and their availability causes a significant threats and challenges. Furthermore, drones supplied with camera or other sensors can threaten the privacy as well as  perform attacks \cite{b3,b4}. For example, on January 8, 2021, A flying drone carrying mobile phones and drugs was discovered near a maximum security prison in Laois, Ireland \cite{1}. Another incident happened on November 24, 2020 a hobbyist drone was loaded with explosives used for bombing in Afghanistan \cite{2}. Also, in Hamburg, Germany 10 malicious drones were spotted in the airport \cite{3}. A recent incident occurred in February, 2021 a drone was equipped with explosives targeting Abaha’s airport was intercepted and destroyed\cite{4}. For that, improving detection methods of unauthorized drones is of great importance \cite{61,62,63,64}. As the consequences of using drones improperly could jeopardize innocents lives.

To prevent this as well as ensure the use of authorized types of drones, many sensors can be used to detect the existence of it. From the these sensors we can find radars, acoustic detector, and cameras. The image-based detection become one of the accurate techniques due to the development of high resolution cameras and computer vision algorithms. These algorithms especially those for object detection can be adopted to detect drone with various types and from different field of view (FOV). Nowadays, deep learning  algorithms are the most accurate and can be used to detect any type of objects with a high performance.From these method we can find deep learning models including YOLOv3, YOLOv4, YOLOv5 and Detectronv2. But the use of these model depend on the purpose of detection as well as the size of dataset used for training each model. for drone detection we can find some size dataset with different types. while the existing methods are used for detecting drone without specifying the type of drone. In addition, the the size of these dataset is small which make the detection performance limited and not accurate for certain scenarios.

In order to overcome these challenges, we provide in this paper a large scale dataset of drone types with different scale and from different FOVs. In addition, we are proposing an evaluation of image-based drone detection using various deep learning models. The open source datasets are discussed in detail as dataset gathering, splitting and configuration. The dataset is collected from different sources. The videos of each class are analysed and the annotations using bounding boxes are performed manually. The proposed dataset provides more details as it is the only dataset that indicates the drone type and the number of images of each type. Also, the training setup was discussed, and some figures of the training results are illustrated. Moreover, the evaluation of each model on the datasets are provided and compared using different metrics including mAP, Precision, Recall and F1-Score.

The rest of this paper will be conducted as follow: section 2 related works, section 3 presents our proposed methods. Experimental results will be shown in section 4, and finally the conclusions will be presented in section 5.

\section{Related works}
A challenging computer vision task is object detection as it involves recognition of the object, predicting the object’s location and it also requires a classification of the detected objects. The section below provides an overview of the existing object detection and classification technologies used in the state of art literature.

The Convolutional Neural Networks models are a very popular models that are used for detection and classification purposes, different images are fed to the network to obtain promising results, and some sources of the images can be a spectrum sensing data \cite{5}, open-source datasets, or manually collected images. The dataset can contain various objects as drones, birds or background \cite{6}, \cite{7}. However, the authors suggested improving the results by including a birds images in the dataset to decrease the false positive in \cite{8}. Some papers combined several CNN models to overcome the existing limitations for the previous proposed methods as inception with FR-CNN, ResNet-101 and the Single Shot Detector SSD-model \cite{9}. As in paper \cite{10}, FR-CNN model is used to overcome the shortcomings, such as the delay of transmitting the image to ground station and the required bandwidth, the authors presented a drone-mounted system. However, to get better results a multi-scale feature fusion is used with a CNN-based model that is applied to detect the small unmanned aerial vehicles \cite{11}. As the Faster R-CNN model is a low-cost and low-power consumption method, the model in this paper \cite{12} was trained on different drone images along with negative objects in the street, such as free-smocking signs, lamps, and trees. Since the Mask R-CNN is an improved algorithm of Faster R-CNN, this paper \cite{b13} presented a detection and segmentation of drone presence in videos using Mask R-CNN. However, in \cite{14} a Pan-Tilt-Zoom camera is used to detect the small drones using six different object detectors and shows a comparison of the obtained results by their accuracy and speed.

In order to detect the drone presence in a video, the authors in \cite{15} decided to use the existing city network monitoring to get the requred data for traning the YOLOv3 model. Also, considering both conditions of day and night in an urban background environment a YOLOv4 model were used  for detecting and localizing drone in \cite{16}. A modified YOLO model are used to get better results for detecting small unmanned aerial vehicles with huge amount of training images used to feed the model \cite{17}, \cite{18}. Also, the authors in \cite{19} and \cite{20}  used YOLOv3 to develop the drone detection system. While in \cite{19}  the tiny YOLOv3  method used exploited. The performance of any algorithm of drone detection can be limited  by the environment changes, the scale if drone as well as the distance between the sensors and the flitting drone \cite{21}. To overcome these challenges, 
other detection methods can be used like radar, while this technique is capable of preforming during all weather conditions. Considering that drones are a low-velocity aircraft; the standard radars are not capable to detect them.  For that, 
 in \cite{22} the X-band radar was used. The X-band radar is affordable and reliable in detecting the UAVs. However, for detecting and tracking the micro-drones, the authors in \cite{23} suggested using the joint range-Doppler-azimuth processing method, after collecting the data from a simple dual-channel Doppler radar. Aslo, in \cite{24} the authors used Passive Bi-static Radar (PBR) system. For testing, two drones were designed to examine the performance of the system, the results were promising. In the same context, the authors in \cite{25} used the micro-Doppler radar for detecting and classifying the small drones, a classifier was used to enhance the detection and distinguish between birds and drones.

\begin{figure*}[h!t]
  \centering
  \footnotesize
    \begin{tabular}[b]{c}
    \includegraphics[width=1\linewidth]{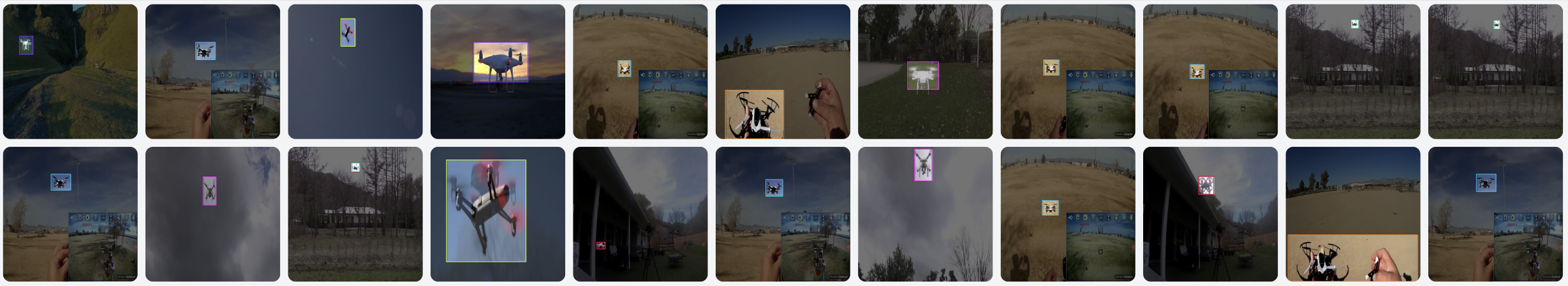} \\
  \end{tabular}  

  \caption{ Drone-type dataset With annotation of different scale.}
  \label{datamodel}
\end{figure*}

\begin{table*}[t!]

    \begin{center}
       \caption{Summary of existing drone detection Datasets } 
       \label{tab1}
       \begin{tabular}{|c|c|c|c|c|c|c|c|c|}
             \hline
        \textbf{Datasets} & \textbf{Size} & \textbf{Resolution} & \textbf{No. of drone types}&\textbf{Annotated Types} &\textbf{Dataset Type}& \textbf{Format}\\
       \hline
        \textbf{Dataset D1} \cite{b32} & 1359 & - &1&0&Image &JPG\\
        \hline
        \textbf{Dataset D2}  \cite{b33} & 4000& - & 1&0& Image&JPEG\\
         \hline
           \textbf{Dataset D3}\cite{b34}  & 2000 &640x512 & 3 &0&Video&MP4\\
               \hline
         \textbf{Our Dataset } & 7000 &1280x720&7&7&Image/Video & JPG \\ 
       \hline
  \end{tabular}
   \end{center}
  \end{table*}

\section{Proposed method}

In order to use a deep learning model for drone detection and classification we proposed a new dataset of drone images with different types. The collected images are annotated also with different formats to fit the process of training the existing methods. The proposed dataset as well as the existing ones are trained and evaluated using existing  deep-learning models including YOLOv3, YOLOv4, YOLOv5, and Detectronv2. In this section, a description of the proposed dataset as well as the deep learning models is provided.

\subsection{Drone-type Dataset}
  
 Drone detection is one of the  challenging tasks in computer vision that can be used to ensure security, prevent attacks, and many other applications. For detecting drones, a set of datasets that are few in the literature are proposed to be used for training deep learning models. The first dataset \textbf{D1} proposed in \cite{b32} consists of 1359 images. The images were collected from Yandex’s image search toolbox, Google, and YouTube. The dataset contains different drone types but the types are not annotated, along with some noisy images. However, the ratio of drone type or noise is not mentioned. On the other hand, the second dataset \textbf{D2} proposed in\cite{b33} is composed of 4000 images. This dataset contains drone-like noise images and non-drone noise images. The images were collected from Google and YouTube. Also, in this dataset, the ratio of drone type and noise is not specified. The third dataset \textbf{D3} \cite{b34} contains two types of recorded videos IR-CAM videos and V-CAM videos. The D3-a consists of images extracted from the IR CAM videos. However, the D3-b consists of images extracted from V-CAM videos, both datasets included four classes, airplanes, birds, drones, and helicopters. The drone class included 2000 images along with 1000 images of helicopters, 1000 bird images, and 1000 airplane images. In addition, the drone class consists of 3 types of drones, the Hubsan H107D+, the high-performance DJI Phantom 4 Pro, and the medium-sized kit drone DJI Flame Wheel. The proposed datasets contain images with their corresponding annotations.

\textbf{ Collected Dataset: }
Drone-type dataset is the first dataset for drone detection with drone images and their types. Besides the collection of drone images with different scales and from different fields of view (FOV) as illustrated in Figure \ref{datamodel}, the proposed dataset allows detection and recognition of the type of flying drone. Our dataset was collected from the Internet including YouTube videos of seven different drone types Bebop, $DJI-phantom-3$, $DJI-phantom-4-pro$, $Emax$, $Intel-Aero-RTF4$, $Mambo$, and $YH-19HW$. The total number of images is 7000 as the Bebop has 1000 images,  $DJI-phantom-3$ has 1000 images, $DJI-phantom-4-Pro$ has 1000 images, $Emax$ has 1000 images, $Intel-Aero-RTF$ has 1000 images, $Mambo$ has 1000 images, and $YH-19HW$ has 1000 images. The description of the proposed dataset and the existing drone detection datasets is given in Table \ref{tab1}.

\subsection{Drone detection and recognition models}
For a better performance, a deep learning model is advised to be trained on a large number of images. The performance of the model can be tested once the model runs on unseen data. Testing the performance of the object detector model can be done by retrieving the metrics as mean average precision (mAP), precision, recall, and F1-score. In this paper, we used the most popular deep learning model for object detection to detect and recognize the drones on the state-of-the-art datasets as well as on the proposed dataset. a description of each one of these models is presented as follows:

\textbf{YOLOv3: }
this model is an improvement of previous versions of YOLOs. It is one of the best models used for real-time processing object detection \cite{26}. The YOLOv3 uses a single neural network and consists of two main components of multi-scale, feature extractor, and detector \cite{27}. The feature extraction is performed using DarkNet53 backbone, which consists of 53 layers. The model predicts the location of the object by a boundary boxes and classify the object.

\textbf{YOLOv4: }
is a developed version of YOLO, while the CSPDarknet53 was used as the new architecture to enhance the learning capability of CNN. Unlike YOLOv3 which was using the FPN, YOLOv4 uses the PANet as a method of parameter aggregation for different levels of detection \cite{b28}. The results of YOLOv4 compared to YOLOv3 improved by increasing the Average Precision (AP) by 10 and the Frame Per Second (FPS) by 12  \cite{b29}.

\textbf{YOLOv5:}
is a single-stage detector and has three important parts: model backbone, model neck and model head. For the model backbone, a cross-stage partial network CSP is used to extract features. The model neck used in YOLOv5 is PANet, to generate feature pyramids which help in scaling the object. A similar model head which was used in YOLOv3 and YOLOv4 is applied in YOLOv5, to perform the final detection that generates output with bounding boxes \cite{b30}.

\textbf{Detectronv2: }
 is the implementation of state-of-art object detection algorithms and it is the next-generation software system of the Facebook AI Research (FAIR). Also, Detectronv2 is an enhanced version of Detectron. Unlike the YOLOs algorithms, Detectronv2 provides an easy API in order to extract the scoring results. It originated from Mask R-CNN and other new features such as Panoptic segmentation \cite{31}. The main goal of Detectron was to offer an efficient codebase for object detection.

\begin{table}[t!]
\footnotesize
    \begin{center}
       \caption{Performance comparison of each model on Drone detection datasets  }
       \label{tab2}
       \begin{tabular}{|c|c|c|c|c|c|}
       \hline
        \textbf{Datasets}&\textbf{Method} & \textbf{mAP} & \textbf{Prec }& \textbf{Recall}& \textbf{F1-Score}\\
        \hline
         D1 \cite{b32}  & YOLOv3& 86.0 & - & - & -\\ %
        \cline{2-6}
        & YOLOv4  & 90.7 & \textbf{90.0 }& 87.0 &\textbf{ 89.0} \\
         \cline{2-6}
          & YOLOv5 &  92.9 & 86.3 & \textbf{91.7} & 88.9 \\
         \cline{2-6}
         &Detectronv2 &\textbf{96.6}   & -&- &- \\ \hline
         
         D2 \cite{b33} &  YOLOv3  & 87.0 & \textbf{98.0} & 86.0 & 87.0 \\
         \cline{2-6}
        & YOLOv4 &\textbf{ 93.5} & 93.0 & \textbf{92.0} & \textbf{93.0 }\\
         \cline{2-6}
        & YOLOv5 & 90.3 & 77.2 & 91.9 & 83.9 \\
         \cline{2-6}
        & Detectronv2  & 84.6  & -&- &- \\  \hline
        
       D3-a \cite{b34}&  YOLOv3  &70.9& \textbf{99.8}& 11.2&  76.0 \\\cline{2-6} 
                             & { YOLOv4 } & {68.7}& {54.0}&{80.0}& { 65.0} \\\cline{2-6}
                             &  { YOLOv5 } & {\textbf{99.4}}&{94.1}&{\textbf{99.4 }}& {\textbf{96.7}} \\\cline{2-6}
                                 & {Detectronv2} & {97.6}&-&-&- \\\cline{2-6} 
        \hline
         
        D3-b \cite{b34}&  YOLOv3 & {72.6 }& {-}& {-}&  {78.4} \\\cline{2-6} 
                             &  { YOLOv5 } &{\textbf{95.3}}&{\textbf{92.8}}& {\textbf{95.1} }& {\textbf{94.0}} \\\cline{2-6}
                                 &{Detectronv2} & {92.0} &-&-&-\\\cline{2-6} 
       
      \hline
       \textbf{Drone-Type} & YOLOv3&  \textbf{99.5}& \textbf{100} & \textbf{99.0} & \textbf{99.0}\\
        \cline{2-6}
         \textbf{(Our)}&YOLOv4  &73.9  &58.0  & 80.0  &  67.0\\
        \cline{2-6}
          & YOLOv5 &  95.3 & 84.9 & 95.4 & 89.8 \\
         \cline{2-6}
         & Detectronv2 & 92.8  &- & -& -\\ \hline
  \end{tabular}
   \end{center}
  \end{table}

\section{Experimental results}

In order to evaluate the proposed dataset using different deep-learning models including YOLO3, YOLO4, YOLO5, and DetectronV2, a set of metrics has been taken into consideration such as MAP, Recall, Precision, F1-measure, and confusion metric. The evaluation has been performed on the exiting methods using the same metrics.


\subsection{Dataset configuration}
In order to train the proposed dataset using different deep learning models, each model requires a specific annotation format. For all the YOLO models each image is resized to 416x416, and the required form of annotations is a .txt file. The annotation file contains the details of the object as well AS the class of the object and the coordinates of the object's location in each image. Unlike the YOLOs, the Detectronv2 annotations follow.JSON format, which includes the boundary box information of the object. 

\subsection{Training setup} All YOLOs models are trained with the same parameter’s configuration. For the split of each dataset,  we split the datasets with 70 for the training set, 10 for the validation set, and 20 for the testing set. As the training process requires a reliable source to train each dataset; all the training was done using Google Colaboratory or “ Google Colab”, it is a cloud version of Python offered by Google.


\subsection{Evaluation Metrics}

In order to have a fast and accurate model for real-time object detection, it is essential to measure the performance of the model and compare the model with certain metrics. mAP and FPS are used here to choose the optimal model for drone detection.


\begin{figure}[t!]
  \centering
  \footnotesize
  \begin{tabular}[b]{c}
    \includegraphics[width=1\linewidth]{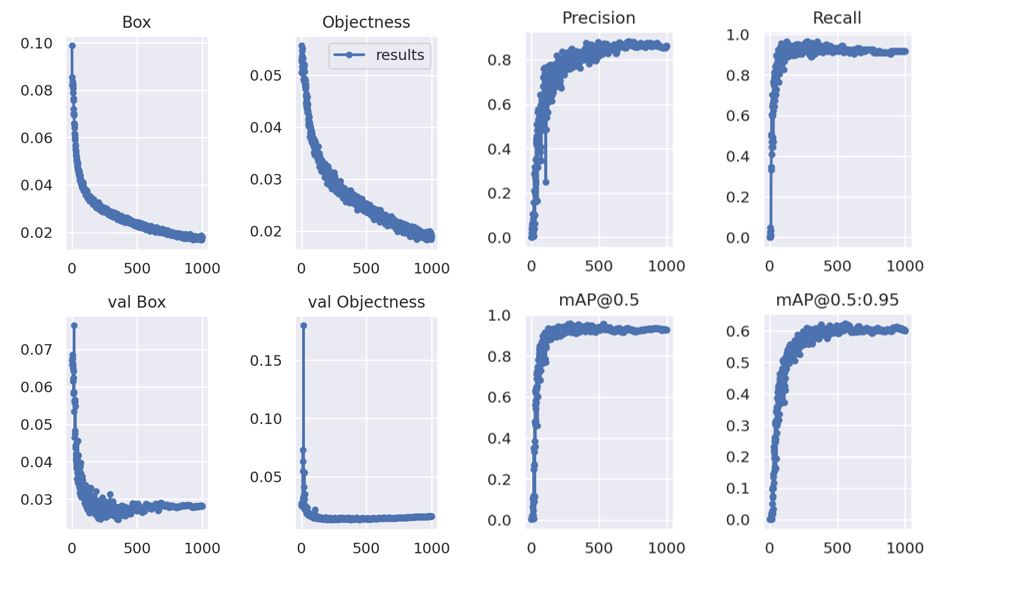} 
  \end{tabular} 
  \caption{ Training outputs of YOLOv5 on Dataset 2.}
  \label{d2}
\end{figure}
\textbf{Precision and Recall:}
Precision is the percentage of the time your prediction is correct. Recall measures how well all the positive predictions are found. Equation (5), and Equation (6) are used to calculate precision and recall \cite{266}.

\begin{equation}
Precision = \frac{TP}{TP+FP}
\end{equation}
\begin{equation}
Recall = \frac{TP}{TP+FN}
\end{equation}

\textbf{Average Precision (AP):}
Precision is the percentage of the time your prediction is correct. Recall measures how well all the positive predictions are found. Equation (5), and Equation (6) are used to calculate precision and recall \cite{266}. 
While mAP, or mean Average Precision, is a metric used to determine performance on various object detection models. Here, the classification and localization of the image are determined \cite{266}. The ground truth of object detection models, the class of the objects, and the bounding box of each object serve as parameters for calculating the mAP. 


\subsection{Evaluation and discussion}
To evaluate the trained networks, the final step is to feed the networks the testing set and check their performance, whether the model is able to correctly detect and classify the object. Moreover, the metrics of each model are obtained, like, precision and recall. The precision is calculated to measure the accuracy of the prediction \cite{35}. However, the recall measures the ability of the model to detect the positive images \cite{35}. In this section, we performed a comparison of the trained models on each dataset including the proposed one.

\textbf{Evaluation on dataset D1:}

\begin{figure}[t!]
  \centering
  \footnotesize
  \begin{tabular}[b]{c}
    \includegraphics[width=1\linewidth]{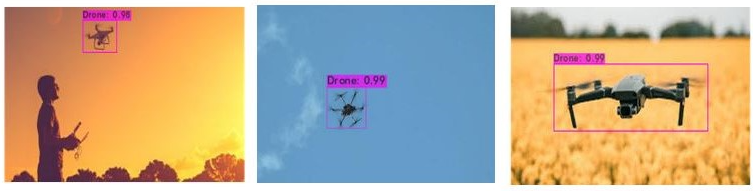} 
  \end{tabular} 
    \begin{tabular}[b]{c}
    \includegraphics[width=1\linewidth]{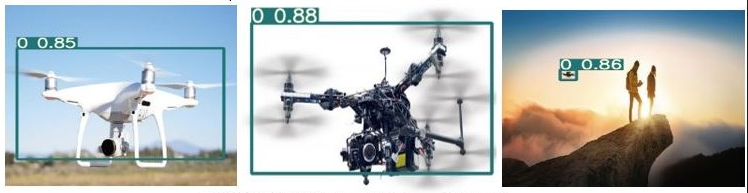} 
  \end{tabular} 
    \begin{tabular}[b]{c}
    \includegraphics[width=1\linewidth]{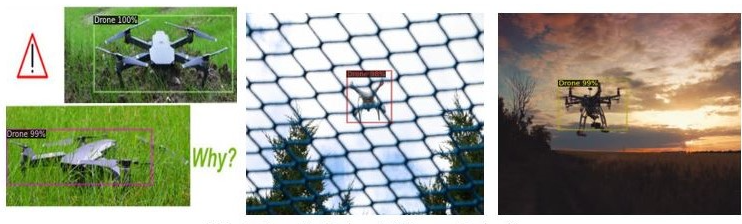} 
  \end{tabular} 
  \caption{Drone detection results on Dataset D1. (First row) YOLOv4. (Second row) YOlOv5. 
 (Third row) Detectronv2.}
  \label{d1vis}
\end{figure}

The first dataset D1, as presented in table \ref{tab2}, provides a comparison of YOLOv3, YOLOv4, YOLOv5, and Detectronv2 results on four datasets including \cite{b32}, \cite{b33}, \cite{b34},and the proposed Drone-Type dataset. As shown, the results of YOLOv4 and YOLOv5 provided higher accuracy compared to the accuracy obtained from tiny YOLOv3 in \cite{b32}. In addition, other metrics are provided such as precision, recall, and F1-score. Furthermore, for the first dataset, the mAP obtained using YOLOv3 is 86 at 2500, while mAP reached 90.7 using YOLOv4. On the other hand, the result using YOLOv5 is 92.9, which is better that YOLOv3 and YOLOv4. The highest mAP is reached using Detectronv2 that better than YOLOv5 by 4\%. 
However, the obtained testing results of the models YOLOv4, YOLOv5, and Detectronv2 are shown in Figure \ref{d1vis}.

\textbf{Evaluation on dataset D2:}

\begin{figure}[t!]
  \centering
  \footnotesize
  \begin{tabular}[b]{c}
    \includegraphics[width=1\linewidth]{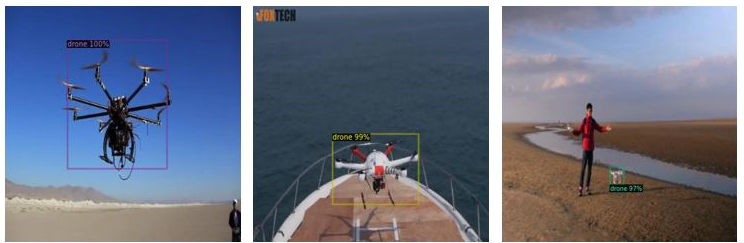} 
  \end{tabular} 
  \caption{Drone detection results on Dataset D2 using YOLOv4 and Detectronv2.}
  \label{d2vis}
\end{figure}

The same deep learning models including YOLOv3, YOLOv4, YOLOv5, and Detectronv2 trained and tested on the second dataset D2 \cite{b33}. The obtained results using object detection metrics are shown in Table \ref{tab2}. From the table, we can observe that YOLOv4 reached the highest mAP, Recal, and F1-score with a value of 93.5, 92, and 93 \% respectively. The second-best result is obtained by YOLOv5 with an mAP of 90.3\% with a difference of 3\% with YOLOv4 and 3\% better than YOLOv3. Using Detectrone we can find that the obtained mAP is the lowest. But in conclusion, we can find that the four models reached close results in terms of all metrics. In addition to the evaluation using metrics, we present also the metrics during the training process using YOLOv5 in Figure \ref{d2}. From this figure, we can find see the evolution of the learning during the training in terms of mAP, Recal, and Precision. For the quantitative representation, we illustrate some drone detection results on dataset D2 in Figure \ref{d2vis}. From the Figure, we can see that the models succeed to detect drones with different scales and sizes as well as outdoor and indoor. 

\begin{figure}[t!]
  \centering
  \footnotesize
  \begin{tabular}[b]{c}
    \includegraphics[width=.41\linewidth]{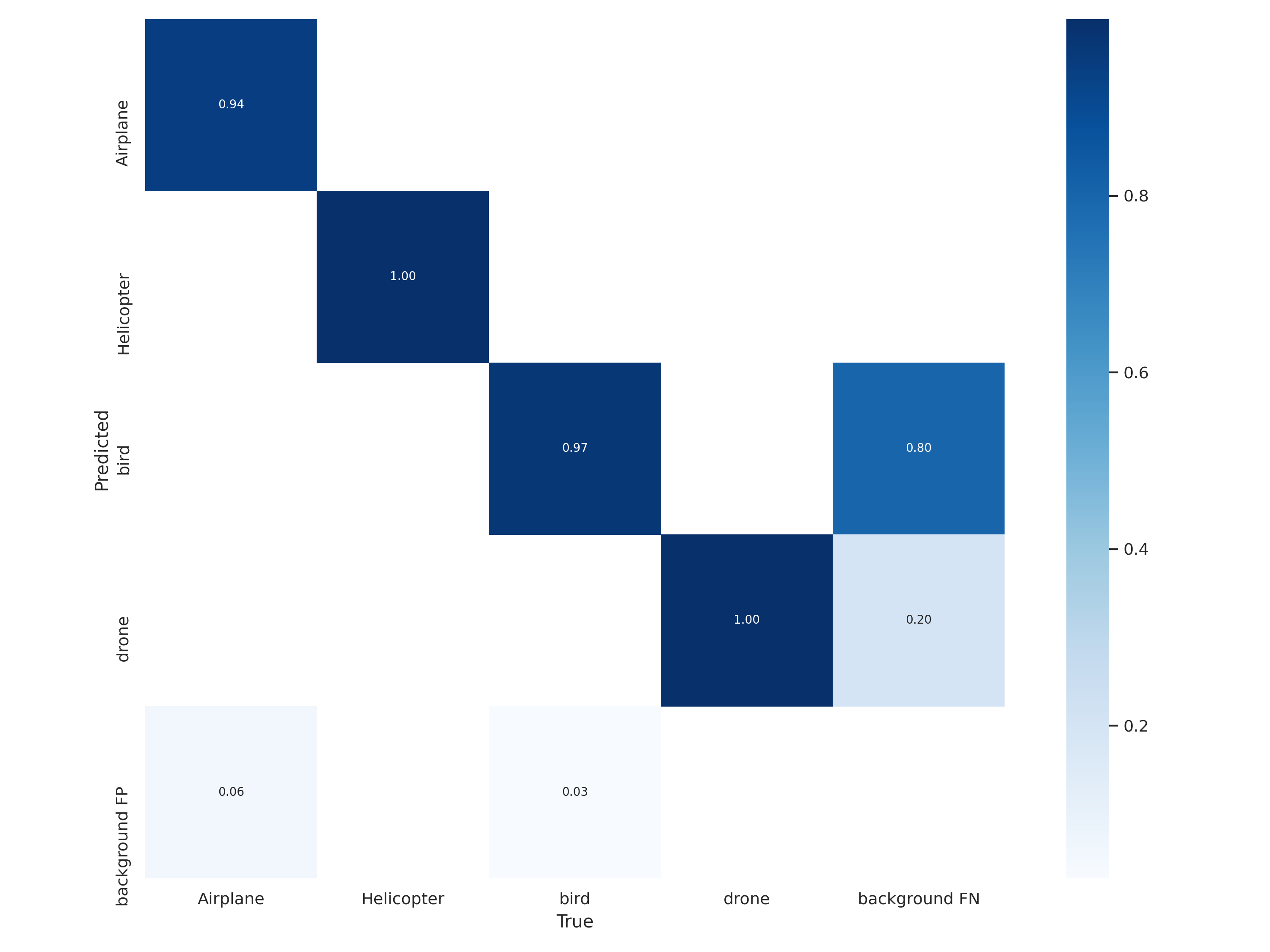} \\
    (a)
  \end{tabular} 
  \begin{tabular}[b]{c}
    \includegraphics[width=.41\linewidth]{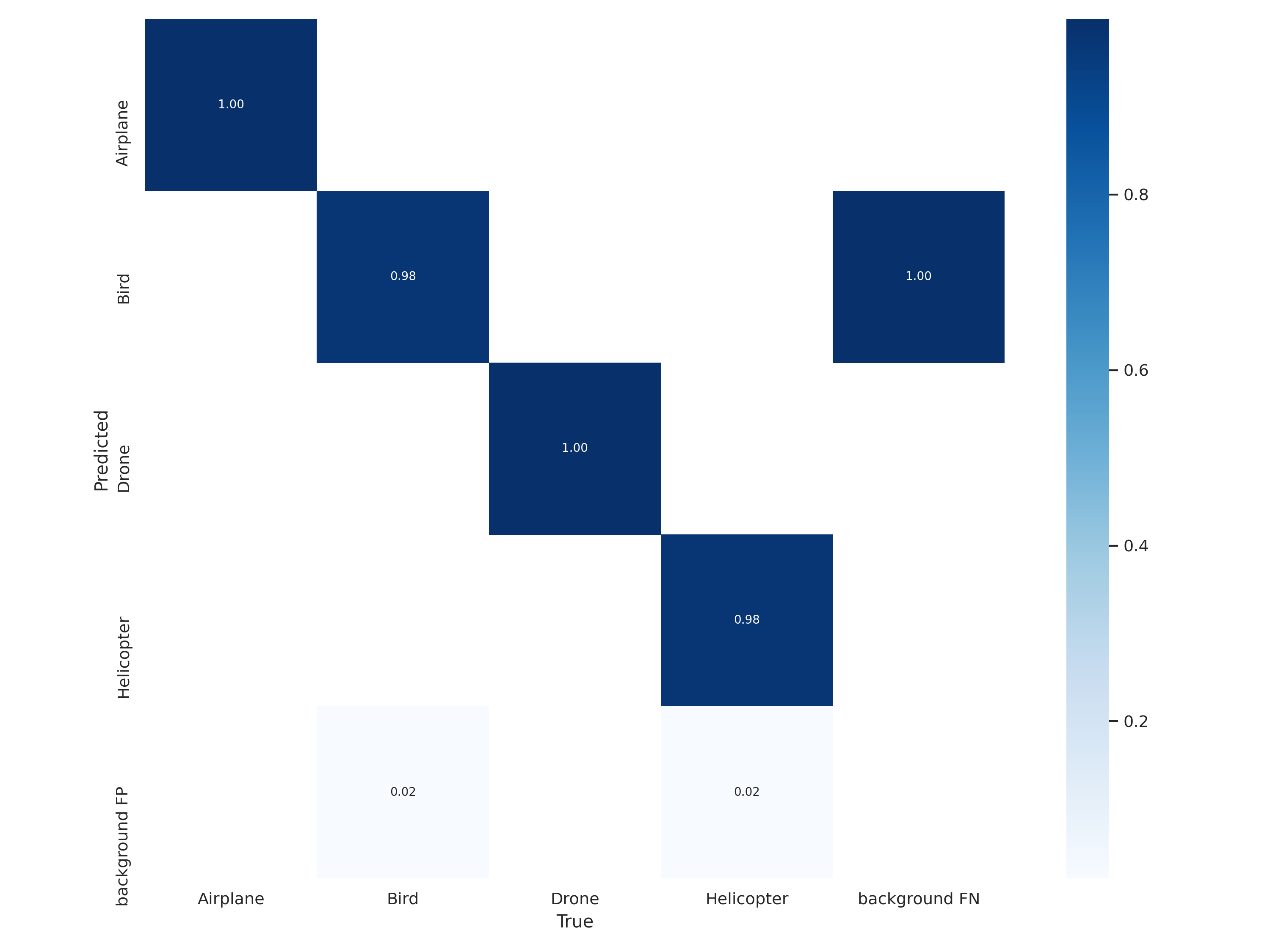} \\
    (b)
  \end{tabular}
  \caption{ (a) YOLOv5: confusion matrix of D3-a. (b) YOLOv5: confusion matrix of D3-b}
  \label{d3}
\end{figure}

\begin{figure}[t!]
  \centering
  \footnotesize
  \begin{tabular}[b]{c}
    \includegraphics[width=1\linewidth]{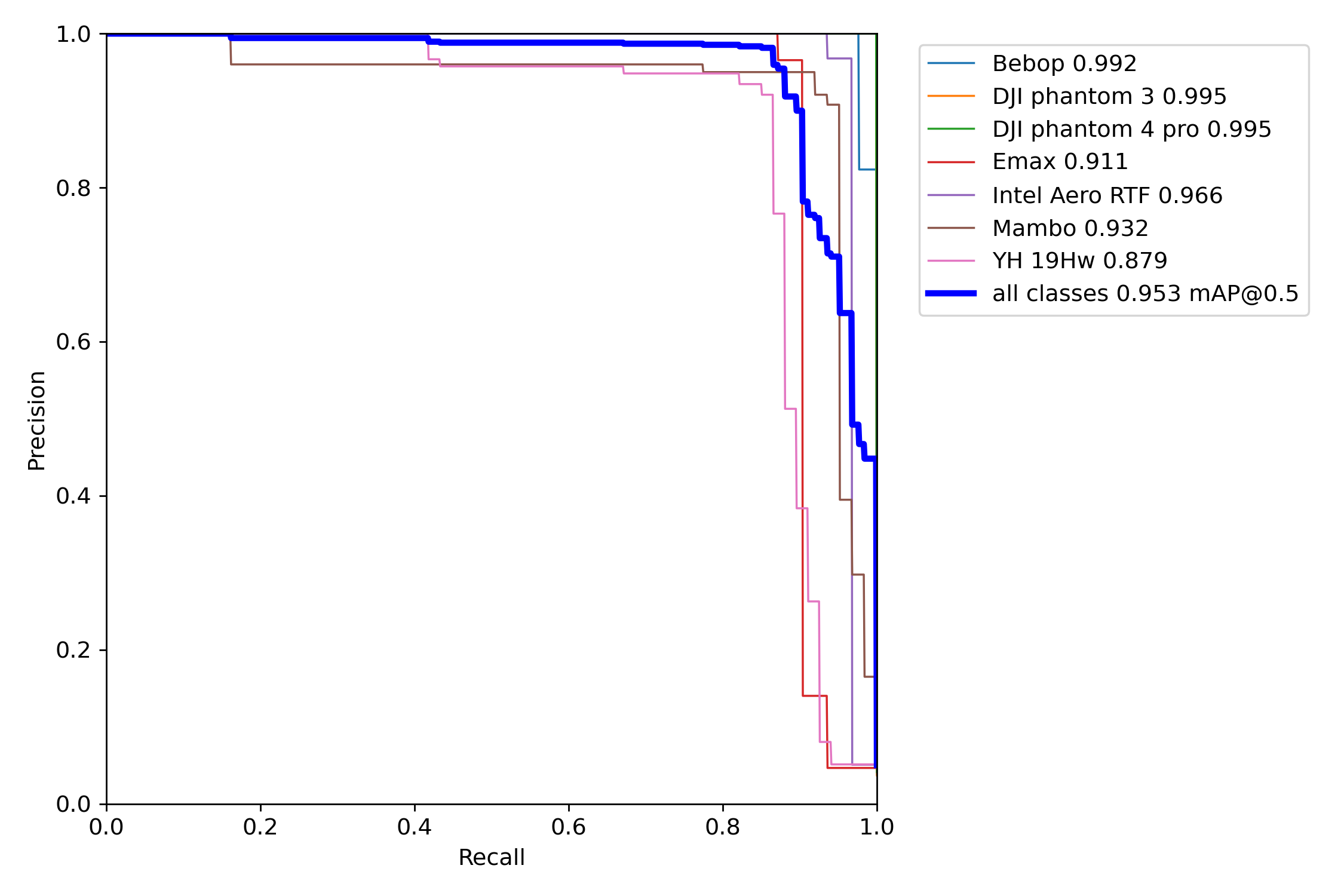} 
  \end{tabular} 
  \caption{Precision-recall curve for propsoed Drone-Type dataset Using YOLOv5}
  \label{dour}
\end{figure}  

\textbf{Evaluation on dataset D3:}

\begin{figure}[t!]
  \centering
  \footnotesize
  \begin{tabular}[b]{c}
    \includegraphics[width=1\linewidth]{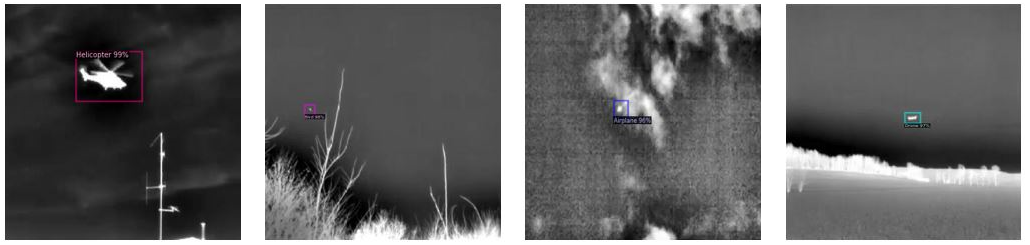} 
  \end{tabular} 
    \begin{tabular}[b]{c}
    \includegraphics[width=1\linewidth]{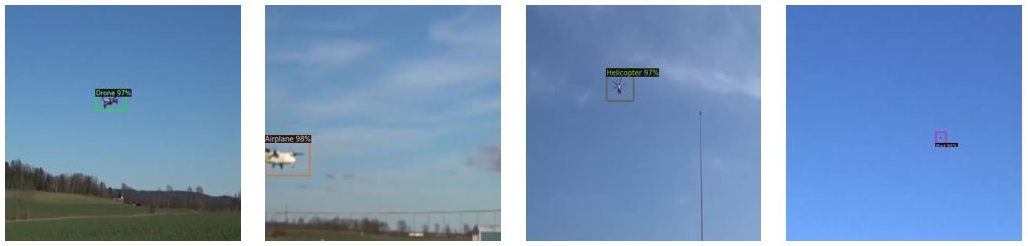} 
  \end{tabular} 
  \caption{Drone detection results on Dataset D3-a and D3-b using YOLOv5 and Detectronv2. }
  \label{d3vis}
\end{figure}

The third dataset \cite{b34} contained two datasets collected using two types of cameras an infrared camera (IR-CAM) and the second one using visible camera (V-CAM). The two datasets contained also different types of airplanes including drones are trained separately. The confusion matrices of the two datasets D3-a and D3-b using YOLOv5 are shown in Figure \ref{d3}. Also obtained evaluation metrics values are shown in Table \ref{tab2}. From the figure \ref{d3} (a), the predicted result of the Airplane was 0.94 and the False Positive (FP) was 0.06. Also, the Bird's predicted results is 0.97 and 0.03 was falsely positive. On the other hand, the results of both Helicopter and Drone were 1.00. However, \ref{d3} (b) shows the results of D3-b, the Airplane, and Drone predicted results were 1.00, and the Bird and Helicopter were 0.98 with 0.02 False Positive (FP) for each class, while the background FN class is falsely detected. These results can be demonstrated by the results presented in table \ref{tab2}, while we can find that the results using metrics on D3-a are better than those on D3-b. For example, for D3-a the result using YOLOv3 from the original paper \cite{b34} of mAP is 70.97 and 72.6 for D3-b. However, the results from YOLOv4, YOLOv5, and Detectronv2 were higher than the accuracy of YOLOv3, while YOLOv5 reached the highest value of most metrics including mAP reached 99.4 for D3-a and 95.3 for D3-b. The visualization of object detection results is presented in Figure \ref{d3vis}. From the figure we can find that the YOLOv5 and Detectronv2 succeed to detect different categories with high performance on infrared and RGB images. Also, these methods can detect the object even if the scale of the object is very small.
    
\begin{table}[t!]
    \begin{center}
       \caption{Details of each class mAP@0.5 of our dataset } 
       \label{tab3}
       \begin{tabular}{|c|c|c|c|}
             \hline
        \textbf{Type of Drone} & \textbf{YOLOv3}&\textbf{YOLOv5} & \textbf{Detectronv2}   \\
       \hline
        Bebop & 99.71 &99.2&64.088\\
        \hline
        DJI phantom 3 & 98.37 &99.5&51.925  \\
         \hline
          DJI phantom 4 Pro &100&99.5&57.184 \\
         \hline
         Emax &100   &91.1&47.217\\ 
       \hline
        Intel Aero RTF &100  &96.6&47.944\\ 
       \hline
        Mambo &99.37  &93.2&64.905  \\ 
       \hline
        YH 19Hw & 99.68& 87.9&38.869   \\ 
       \hline
  \end{tabular}
   \end{center}
  \end{table}

\begin{figure}[t!]
  \centering
  \footnotesize
  \begin{tabular}[b]{c}
    \includegraphics[width=1\linewidth]{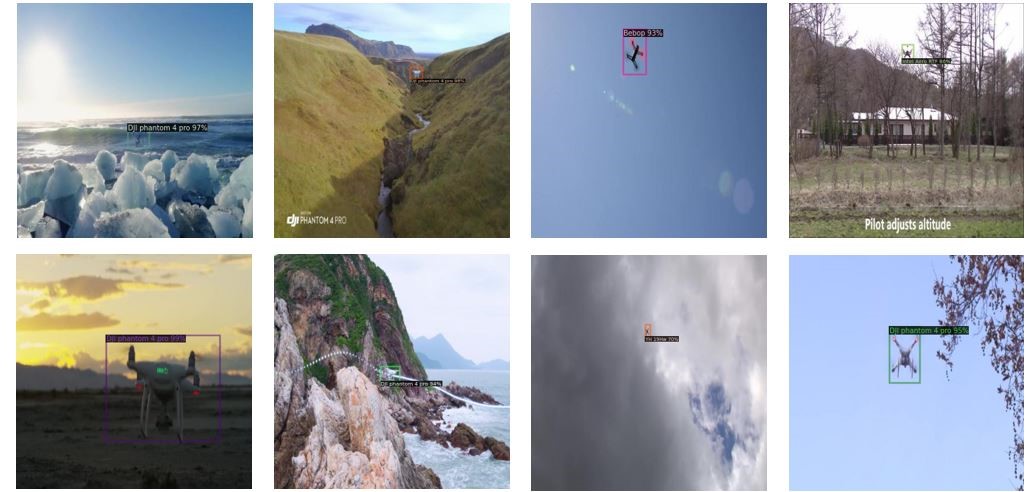} \\

  \end{tabular} 

  \caption{Drone detection results on Drone-Type  dataset using YOLOv3  and YOLOv5.}
  \label{ourvis}
\end{figure}

\textbf{Evaluation on the proposed Drone-Type dataset:}
Using the same configuration as well as the presentation of the obtained quantitative and qualitative results for the three datasets, we performed an evaluation of the deep learning method on the proposed Drone-Type dataset. In table \ref{tab2} the obtained results on the Drone-type dataset using the four deep learning methods, shows that YOLOv3 is the best method for detecting drone in terms of all metrics including mAP of 99.5.
While the lowest mAP@0.5 was the YOLOv4 with 73.94. On the other hand, for YOLOv5, the average mAP of the 7 classes was 95.3, and using the Detectronv2 scored 92.8. 
This can be shown also in Table \ref{tab3} which represents the obtained mAP values for each class using YOLOv3, YOLOv5, and Detectronv2. The highest mAP score obtained using YOLOv3 is 99.7 for Bebop type, 100 for DJI phantom 4 Pro type as well as Emax and Intel Aero RTF, but the YH 19HW was 99.68. We can notice that Detectronv2 gives the lowest mAP values compared to YOLOv3 and YOLOv5. The details of each class along with their average mAP  using YOLOv5 are plotted in Figure \ref{dour}. Visually, the obtained results using YOLOv3 and YOlOv5 of some samples from the Drone-Type dataset are illustrated in Figure \ref{ourvis}. From the visualized results we can see that the methods can detect the drones with their classes even if the images contain objects of different scales. Unlike the other datasets, our dataset can give the type of drones  as well as the percentage of a drone with this type. This can be used for many other applications for precise detection.

 \section{Conclusion}

This paper presents the implementation of the deep learning models YOLOv3, YOLOv4, YOLOv5, and Detectronv2 used for drone detection. A comparison to state of art literature using various datasets and an analysis of experiments were shown. In terms of mAP efficiency, the best result for the first dataset was obtained using Detectronv2 and provided 96.672. However, for the second dataset YOLOv4 presented the best result of 93.57. On the other hand, the third dataset highest accuracy was obtained using YOLOv5 for D3-a was 99.49 and 95.36 for D3-b. Finally, for our dataset the YOLOv3 provided the highest accuracy 99.58.

\section*{Acknowledgment}

This research work was made possible by research grant support (QUEX-CENG-SCDL-19/20-1 ) from Supreme Committee for Delivery and Legacy (SC) in Qatar.


\begin{thebibliography}{00}
\bibitem{b1}Himeur, Y., Al-Maadeed, S., Almaadeed, N., Abualsaud, K., Mohamed, A., Khattab, T., \& Elharrouss, O. (2022). Deep visual social distancing monitoring to combat COVID-19: A comprehensive survey. Sustainable cities and society, 85, 104064.

\bibitem{b2}Ottakath, N., Elharrouss, O., Almaadeed, N., Al-Maadeed, S., Mohamed, A., Khattab, T., \& Abualsaud, K. (2022). ViDMASK dataset for face mask detection with social distance measurement. Displays, 73, 102235.

\bibitem{b3}Abbad, A., Elharrouss, O., Abbad, K., \& Tairi, H. (2018). Application of MEEMD in post-processing of dimensionality reduction methods for face recognition. Iet Biometrics, 8(1), 59-68.

\bibitem{b4}Elharrouss, O., Almaadeed, N., \& Al-Maadeed, S. (2020, February). An image steganography approach based on k-least significant bits (k-LSB). In 2020 IEEE International Conference on Informatics, IoT, and Enabling Technologies (ICIoT) (pp. 131-135). IEEE.

\bibitem{1}R. O'Connor, “Man arrested for flying drone carrying mobile phones and drugs near maximum security prison in Laois,” The Irish Post, 08-Jan-2021. [Online]. Available:  https://www.irishpost.com/news/man-arrested-for-flying-drone-carrying-mobile-phones-and-drugs-near-maximum-security-prison-in-laois-201104.

\bibitem{2} Krone.at, “Taliban nutzen Hobby-Drohnen als Sprengstoffträger,” Kronen Zeitung, 24-Nov-2020. [Online]. Available: https://www.krone.at/2282810. 

\bibitem{3}“Map of World Wide Drone Incidents,” Dedrone. [Online]. Available: \underline{https://www.dedrone.com/resources/incidents/all.} 

\bibitem{4}
Al Jazeera, “Saudi Arabia says it foiled Houthi drone attack on Abha airport,” Houthis News | Al Jazeera, 13-Feb-2021. [Online]. Available: https://www.aljazeera.com/news/2021/2/13/saudi-arabia-says-it-foiled-houthi-drone-attack-on-abha-airport. 

\bibitem{61}Riahi, A., Elharrouss, O., \& Al-Maadeed, S. (2022). BEMD-3DCNN-based method for COVID-19 detection. Computers in biology and medicine, 142, 105188.


\bibitem{62}Elharrouss, O., Almaadeed, N., Abualsaud, K., Al-Maadeed, S., Al-Ali, A., \& Mohamed, A. (2022). FSC-set: counting, localization of football supporters crowd in the stadiums. IEEE Access, 10, 10445-10459.


\bibitem{63}Elharrouss, O., Almaadeed, N., \& Al-Maadeed, S. (2020, February). LFR face dataset: Left-Front-Right dataset for pose-invariant face recognition in the wild. In 2020 IEEE International Conference on Informatics, IoT, and Enabling Technologies (ICIoT) (pp. 124-130). IEEE.

\bibitem{64}Elharrouss, O., Almaadeed, N., Abualsaud, K., Al-Ali, A., Mohamed, A., Khattab, T., \& Al-Maadeed, S. (2021). Drone-SCNet: Scaled cascade network for crowd counting on drone images. IEEE Transactions on Aerospace and Electronic Systems, 57(6), 3988-4001.

\bibitem{5} H. Chen, Z. Wang and L. Zhang, "Collaborative spectrum sensing for illegal drone detection: A deep learning-based image classification perspective," in China Communications, vol. 17, no. 2, pp. 81-92, Feb. 2020, 

\bibitem{6}H. M. Oh, H. Lee and M. Y. Kim, "Comparing Convolutional Neural Network(CNN) models for machine learning-based drone and bird classification of anti-drone system," 2019 19th International Conference on Control, Automation and Systems (ICCAS), Jeju, Korea (South), 2019, pp. 87-90.

\bibitem{7} Seidaliyeva, U.; Akhmetov, D.; Ilipbayeva, L.; Matson, E.T. Real-Time and Accurate Drone Detection in a Video with a Static Background. \textit{Sensors} \textbf{2020}, 20, 3856.

\bibitem{8} M. Saqib, S. Daud Khan, N. Sharma and M. Blumenstein, "A study on detecting drones using deep convolutional neural networks," 2017 14th IEEE International Conference on Advanced Video and Signal Based Surveillance (AVSS), Lecce, 2017, pp. 1-5.



\bibitem{9} M. Nalamati, A. Kapoor, M. Saqib, N. Sharma and M. Blumenstein, "Drone Detection in Long-Range Surveillance Videos," 2019 16th IEEE International Conference on Advanced Video and Signal Based Surveillance (AVSS), Taipei, Taiwan, 2019, pp. 1-6.

\bibitem{10} C. Wang, R. Zhao, X. Yang and Q. Wu, "Research of UAV target detection and flight control based on deep learning," 2018 International Conference on Artificial Intelligence and Big Data (ICAIBD), Chengdu, 2018, pp. 170-174.

\bibitem{11} Y. Zhang, Y. Zhang, Z. Shi, J. Zhang and M. Wei, "Design and Training of Deep CNN-Based Fast Detector in Infrared SUAV Surveillance System," in IEEE Access, vol. 7, pp. 137365-137377, 2019.

\bibitem{12} I. G. M. I. Moteir, K. Ismail, F. M. Zawawi and M. M. M. Azhar, "Urban Intelligent Navigator for Drone Using Convolutional Neural Network (CNN)," 2019 International Conference on Smart Applications, Communications and Networking (SmartNets), Sharm El Sheik, Egypt, 2019, pp. 1-4.



\bibitem{b13} K. V. V. Subash, M. V. Srinu, M. R. V. Siddhartha, N. C. S. Harsha and P. Akkala, "Object Detection using Ryze Tello Drone with Help of Mask-R-CNN," 2020 2nd International Conference on Innovative Mechanisms for Industry Applications (ICIMIA), Bangalore, India, 2020, pp. 484-490.

\bibitem{14} J. Park, D. H. Kim, Y. S. Shin and S. Lee, "A comparison of convolutional object detectors for real-time drone tracking using a PTZ camera," 2017 17th International Conference on Control, Automation and Systems (ICCAS), Jeju, Korea (South), 2017, pp. 696-699.

\bibitem{15} TAO Lei, HONG Tao, CHAO Xu. Drone identification and location tracking based on YOLOv3[J]. Chinese Journal of Engineering, 2020, 42(4): 463-468.

\bibitem{16} Igor S. Golyak, Dmitriy R. Anfimov, Igor L. Fufurin, Andrey L. Nazolin, Sergey V. Bashkin, Vladimir L. Glushkov, Andrey N. Morozov, "Optical multi-band detection of unmanned aerial vehicles with YOLO v4 convolutional neural network," Proc. SPIE 11525, SPIE Future Sensing Technologies, 115250Y (8 November 2020); 

\bibitem{17} Wu M., Xie W., Shi X., Shao P., Shi Z. (2018) Real-Time Drone Detection Using Deep Learning Approach. In: Meng L., Zhang Y. (eds) Machine Learning and Intelligent Communications. MLICOM 2018. Lecture Notes of the Institute for Computer Sciences, Social Informatics and Telecommunications Engineering, vol 251. Springer, Cham.

\bibitem{18} Liu, M.; Wang, X.; Zhou, A.; Fu, X.; Ma, Y.; Piao, C. UAV-YOLO: Small Object Detection on Unmanned Aerial Vehicle Perspective. Sensors 2020, 20, 2238.

\bibitem{19} N. Souli et al., "HorizonBlock: Implementation of an Autonomous Counter-Drone System," 2020 International Conference on Unmanned Aircraft Systems (ICUAS), Athens, Greece, 2020, pp. 398-404.

\bibitem{20} D. K. Behera and A. Bazil Raj, "Drone Detection and Classification using Deep Learning," 2020 4th International Conference on Intelligent Computing and Control Systems (ICICCS), Madurai, India, 2020, pp. 1012-1016.
\bibitem{21} H. M. Oh, H. Lee and M. Y. Kim, "Comparing Convolutional Neural Network(CNN) models for machine learning-based drone and bird classification of anti-drone system," 2019 19th International Conference on Control, Automation and Systems (ICCAS), Jeju, Korea (South), 2019, pp. 87-90.

\bibitem{22} I. Bouzayene, K. Mabrouk, A. Gharsallah and D. Kholodnyak, "Scan Radar Using an Uniform Rectangular Array for Drone Detection with Low RCS," 2019 IEEE 19th Mediterranean Microwave Symposium (MMS), Hammamet, Tunisia, 2019, pp. 1-4.

\bibitem{23} M. Jian, Z. Lu and V. C. Chen, "Drone detection and tracking based on phase-interferometric Doppler radar," 2018 IEEE Radar Conference (RadarConf18), Oklahoma City, OK, USA, 2018, pp. 1146-1149, doi: 10.1109/RADAR.2018.8378723.

\bibitem{24}Y. Liu, X. Wan, H. Tang, J. Yi, Y. Cheng and X. Zhang, "Digital television based passive bistatic radar system for drone detection," 2017 IEEE Radar Conference (RadarConf), Seattle, WA, 2017, pp. 1493-1497.

\bibitem{25} S. Björklund, "Target Detection and Classification of Small Drones by Boosting on Radar Micro-Doppler," 2018 15th European Radar Conference (EuRAD), Madrid, Spain, 2018, pp. 182-185, doi: 10.23919/EuRAD.2018.8546569.


\bibitem{26} U. Almog "YOLO V3 Explained" Medium, 13-Oct-2020. [Online].
Available: \underline{https://towardsdatascience.com/yolo-v3-explained-ff5b850390f}
\bibitem{27} J. Brownlee, “A Gentle Introduction to Object Recognition with Deep Learning,” Machine Learning Mastery, 05-Jul-2019.  [Online]. Available: \underline {http://CRAN.R-project.org/package=raster} 
\bibitem{b28} R. Orac, “What's new in YOLOv4?” Medium, 13-Dec-2020. [Online].
Available:
\underline{https://cutt.ly/njMLNRW}

\bibitem{b29} M. Rajput, “YOLO V5-Explained and Demystified,” Medium, 01-Jul-2020.  [Online]. Available: \underline{https://cutt.ly/PjMXrmY} 

\bibitem{b30} Y. Liu, “The Confusing Metrics of AP and mAP for Object Detection,” Medium, 18-May-2020. [Online]. Available: \underline{https://cutt.ly/5jMXpMz}

\bibitem{31}R. O'Connor, “Man arrested for flying drone carrying mobile phones and drugs near maximum security prison in Laois,” The Irish Post, 08-Jan-2021. [Online]. Available: https://www.irishpost.com/news/man-arrested-for-flying-drone-carrying-mobile-phones-and-drugs-near-maximum-security-prison-in-laois-201104. 

\bibitem{b32} Dasmehdix, “dasmehdix/drone-dataset,” GitHub.[Online]. Available: \underline{https://github.com/dasmehdix/drone-dataset}

\bibitem{b33} Aksoy, Mehmet Çağrı; Orak, Alp Sezer; Özkan, Hasan Mertcan; Selimoğlu, Bilgin (2019), “Drone Dataset: Amateur Unmanned Air Vehicle Detection”, Mendeley Data, V4, 

\bibitem{b34} Svanström, F. (2020). Drone Detection and Classification using Machine Learning and Sensor Fusion.

\bibitem{35} A. Coluccia et al., "Drone-vs-Bird Detection Challenge at IEEE AVSS2019," 2019 16th IEEE International Conference on Advanced Video and Signal Based Surveillance (AVSS), Taipei, Taiwan, 2019, pp. 1-7. 

\bibitem{266}    R. Padilla, S. L. Netto, and E. A. da Silva, "A survey on performance metrics for object-detection algorithms," in 2020 International Conference on Systems, Signals and Image Processing (IWSSIP), 2020: IEEE, pp. 237-242. 

\bibitem{299}    J. Licheng, F. Zhang, F. Liu, S. Yang, L. Li, Z. Feng, and R. Qu. "A survey of deep learning-based object detection." IEEE Access 7 (2019): 128837-128868.68.
\end{thebibliography}
\end{document}